\documentclass{article}
\pdfoutput=1 
    \PassOptionsToPackage{numbers, square, compress}{natbib}

\usepackage[final]{neurips_2021}

\usepackage[utf8]{inputenc} 
\usepackage[T1]{fontenc}    
\usepackage{hyperref}       
\usepackage{url}            
\usepackage{booktabs}       
\usepackage{amsfonts}       
\usepackage{nicefrac}       
\usepackage{microtype}      
\usepackage{xcolor}         

\usepackage{amsmath}
\usepackage[pdftex]{graphicx}

\title{Risk Sensitive Model-Based Reinforcement Learning using Uncertainty Guided Planning}

\author{
  Stefan Radic Webster\thanks{Code: \href{https://github.com/sradicwebster/mbrl-lib/tree/uncertainty_guided_planning/uncertainty_guided_planning}{https://github.com/sradicwebster/mbrl-lib/tree/uncertainty\_guided\_planning/uncertainty\_guided\_planning}} \\
  Department of Computer Science \\
  University of Bristol \\
  \texttt{s.radicwebster@bristol.ac.uk} \\
\And
   Peter Flach \\
   Department of Computer Science \\
   University of Bristol \\
   \texttt{peter.flach@bristol.ac.uk} \\
}

\begin{document}

\maketitle

\begin{abstract}
  Identifying uncertainty and taking mitigating actions is crucial for safe and trustworthy reinforcement learning agents, especially when deployed in high-risk environments. In this paper, risk sensitivity is promoted in a model-based reinforcement learning algorithm by exploiting the ability of a bootstrap ensemble of dynamics models to estimate environment epistemic uncertainty. We propose uncertainty guided cross-entropy method planning, which penalises action sequences that result in high variance state predictions during model rollouts, guiding the agent to known areas of the state space with low uncertainty. Experiments display the ability for the agent to identify uncertain regions of the state space during planning and to take actions that maintain the agent within high confidence areas, without the requirement of explicit constraints. The result is a reduction in the performance in terms of attaining reward, displaying a trade-off between risk and return.
\end{abstract}

\section{Introduction}
\label{sec:intro}

The ability to internally simulate the consequences of proposed actions using a model of the environment is a key advantage of model-based reinforcement learning (MBRL) algorithms over model-free counterparts in attaining safe behaviour in high-risk domains. Safety is crucial to making progress towards the deployment of real-world RL, although it is rarely the main consideration in the development of MBRL algorithms, with motivations commonly cited as improving sample-efficiency, enhanced exploration via intrinsic motivation and transfer learning to new tasks \cite{moerland_model-based_2020}.

The goal of safe RL is to avoid high-risk or dangerous situations \cite{garcia_comprehensive_2015}. In this paper, unsafe regions are implicitly defined as areas of the state space where there is a lack of knowledge, as identified by dynamics model epistemic uncertainty. The aim is to design an algorithm that displays risk-adverse behaviour upon approaching uncertain situations to ensure the agent remains in high confidence regions, preventing unpredictable and potentially dangerous outcomes. This is achieved by modifying the cross-entropy method (CEM) \cite{rubinstein_cross-entropy_1999} used for action selection in the probabilistic ensembles with trajectory sampling (PETS) \cite{chua_deep_2018} algorithm. By utilising the ability of an ensemble of probabilistic dynamics models to capture both epistemic and aleatoric uncertainty in the environment, the uncertainty guided CEM planner directs the agent toward areas of the state space where confidence in model rollouts are high. Specifically, the objective function is modified to include a penalty term for the variance of state predictions during planning. The idea stems from the risk-sensitive RL objective \cite{heger_consideration_1994}, which penalises the variance in return of a trajectory, and the notion of state controllability as a method for guided exploration based on the predictability of a trajectory \cite{gehring_smart_2013}. The intuition behind uncertainty guided planning is that for safety-critical domains, agents should seek out areas of the state space which are predictable in order to ensure trustworthy behaviour.

\section{Related Work}
\label{sec:related}

Safety in RL has been addressed by modifying the objective function to account for risk due to uncertainty about the environment. Typically, this is done to meet one of a number of goals: i) to maximise return for the worse case scenario \cite{heger_consideration_1994} using the robust  Markov Decision Processes (MDP) formulisation \cite{iyengar_robust_2005}, ii) to promote risk-sensitivity by penalising some metric of risk (frequently the variance of return) \cite{gosavi_reinforcement_2009} or iii) to impose hard constraints (or fix the probability of constraint violation) on hazardous areas of the state space \cite{geibel_risk-sensitive_2005} by modelling the problem as a constrained MDP \cite{altman_asymptotic_1993}. An alternative approach is to prevent exploratory actions causing dangerous scenarios. These safe exploration strategies can utilise external knowledge, such as learning from demonstrations \cite{argall_survey_2009} or teacher advice \cite{clouse_integrating_1996}, or carry out risk directed exploration; one example is the use of an exploration bonus proportional to the predictability of the state \cite{gehring_smart_2013}.

The majority of proposed methods for safe RL are model-free. Approaches to safe MBRL typically utilise models to enforce constraint satisfaction \cite{fisac_general_2019,dalal_safe_2018,chow_lyapunov-based_2018}. Focusing on methods that are based on PETS or use CEM planning, Zhang et al. \cite{zhang_cautious_2020} propose safety-critical adaptation during particle propagation based on two methods for risk aversion: robustness to the worse case return and avoidance of catastrophic states by penalising the probability of encountering high-cost states. Likewise, Thananjeyan et al. \cite{thananjeyan_safety_2020} use a chance constrained MDP formulation but instead discard action sequences if a certain constraint violation threshold is exceeded, which is similar to constrained CEM \cite{wen_constrained_2018}. Liu et al. \cite{liu_constrained_2021} propose robust CEM, where a classifier is learnt as the constraint function and the optimisation is with respect to the worse case scenario over an ensemble of dynamics models. Uncertainty guided CEM, as proposed here, formulates safety as the avoidance of uncertain regions of the state space, which has parallels to offline RL, as opposed to satisfying safety constraints which requires an explicit definition of unsafe states as part of a constrained MDP.

Offline RL addresses the problem of learning policies from previously collected datasets without additional online interaction \cite{levine_offline_2020}. Specific offline MBRL algorithms have been proposed with a risk-sensitive objective, similar to that used in uncertainty guided CEM, as a way to mitigate distributional shift. Kidambi et al. \cite{kidambi_morel_2020} formulate the problem as a pessimistic MDP by the addition of a heavily penalised terminating state for out-of-distribution state-action pairs as identified by the maximum dynamics ensemble discrepancy, whilst Yu et al. \cite{yu_mopo_2020} penalise uncertainty as the maximum variance of ensemble predictions. These offline RL algorithms learn a policy from a static dataset, as opposed to performing online decision time planning over a learnt model. 

\section{Background}
\label{sec:background}

RL problems are typically formulated as a Markov Decision Processes (MDP) \cite{bellman_markovian_1957} consisting of states $s_t \in S$, actions $a_t \in A$, a reward function $r(s_t,a_t)$ and a transition function $p(s_{t+1}|s_t,a_t)$ representing the environment dynamics as the probability of transitioning to state $s_{t+1}$ given action $a_t$ was selected from state $s_t$. The objective is to learn a policy $\pi(a|s)$, which maps states to actions such that the expected cumulative reward is maximised: $\pi^\star = arg\ \underset{\pi}{max}\ \mathbb{E}_\pi  \left[ \sum\nolimits_t r(s_t,a_t) \right] $ \cite{sutton_reinforcement_2018}.

Inherent to model-based RL is planning over a learnt model of the environment dynamics. The model approximates the transition function $s_{t+1} = f_\theta(s_t,a_t)$, with the parameters $\theta$ trained to maximise the log-likelihood of the observed data $\mathcal{D} = \{s_n,a_n,s_{n+1}\}_{n=1}^D$. MBRL algorithms can be broadly grouped into those that perform background planning, which generates imaginary data using the model and updates a policy and/or value function, or decision time planning, which performs online trajectory optimisation from the current state $s_0$ over a finite horizon $H$,
\begin{equation}
    a^\star_{0:H-1} = arg\ \underset{a_{0:H-1}}{max}\ \mathbb{E} \left[ \sum_{t=0}^{H-1} r(s_t,a_t) \right], \quad  s_{t+1} \sim f_\theta(s_t,a_t).
\end{equation}
Selecting $a^\star_0$ after planning from $s_0$ and replanning at the next time step is known as Model Predictive Control (MPC). A popular method for decision time planning is the cross-entropy method (CEM) \cite{rubinstein_cross-entropy_1999}, a population-based stochastic optimisation technique. CEM involves sampling a population with size $N$ of $H$ length action sequences from a planning distribution, commonly a multivariate Gaussian with diagonal covariance matrix for continuous actions, $\mathcal{X}_1,...,\mathcal{X}_N \sim \mathcal{N}(\mu,\Sigma)\ \text{where}\ \mu \in \mathbb{R}^H, \Sigma \in \mathbb{R}^{H\times H} $, simulating the actions using the learnt dynamics model and re-fitting the planning distribution using the elite-set of the highest performing action sequences from the population. This procedure is repeated for a set number of planning iterations.

Two kinds of uncertainty are relevant to the modelling of environment dynamics \cite{chua_deep_2018}. The first of these is aleatoric uncertainty due to inherent environment stochasticity, which is captured by learning a probabilistic model which predicts a distribution over next states, typically a Gaussian distribution in continuous state spaces, $s_{t+1} \sim \mathcal{N}(\mu_\theta(s_t,a_t),\Sigma_\theta(s_t,a_t))$. The other is epistemic, or model, uncertainty due to a lack of knowledge about the environment, which can be captured by using an ensemble of models trained independently by bootstrapping.

The PETS algorithm \cite{chua_deep_2018}, which forms the starting point of this work, performs CEM planning over an ensemble of probabilistic neural network dynamics models. A particle-based state propagation method known as trajectory sampling (TS) is used during planning, where $P$ copies of the current state are made for each of the action sequences in the population and the particles are propagated by sampling a model from the ensemble. For a particular particle the dynamics model can be either uniformly re-sampled from the ensemble every time step (TS1) or fixed for the entire rollout (TS$\infty$).

\section{Uncertainty Guided CEM Planning}
\label{sec:uncertainty}

Uncertainty guided CEM planning utilises the particle propagation methods in PETS to identify uncertainty in state predictions during model rollouts. By sampling a dynamics model from the ensemble for each particle and keeping the model fixed for the rollout (TS$\infty$ state propagation in \cite{chua_deep_2018}), aleatoric and epistemic uncertainties are separated. The epistemic uncertainty of the learned model arises due to insufficient knowledge about the environment dynamics, and is the greater concern to ensure safe and trustworthy behaviour. For a particular action sequence in the population $\mathcal{X}_i$, the epistemic uncertainty $\omega_i$ is estimated as the variance of the state predictions, $\sigma^2_{S,t}$, over $P$ particles for state dimension $S$ after $t$ step rollouts using a model sampled from the ensemble, averaged over the horizon $H$ and across all state dimensions,
\begin{equation}
    \omega_i = \frac{1}{dim(S)} \sum_S \frac1H \sum_{t=1}^H \frac{\sigma^2_{S,t}}{\bar{\sigma}^2_{S,t}},
    \label{eq:uncert}
\end{equation}
$$\sigma^2_{S,t} = \frac1P \sum_{p=1}^P (f_{\theta_p} - \mu_p)^2, \quad \text{where,} \quad \mu_p = \frac1P \sum_{p=1}^P f_{\theta_p}.$$
$\sigma^2_{S,t}$ is normalised by the mean variance for the state dimension $S$ after $t$ step model rollouts, $\bar{\sigma}^2_{S,t}$, calculated from previous model rollouts. This is to account for the higher variance of state predictions as more steps are predicted into the future, and ensures the uncertainty estimation is equally weighted across all time steps in the planning horizon.

Uncertainty estimation using the variance over an ensemble of learnt probabilistic neural networks has been shown to be an effective method to identify discrepancies between the true and learned dynamics \cite{lu_revisiting_2021}. However, this method does not completely decouple epistemic and aleatoric uncertainty\footnote{An estimation of epistemic uncertainty is the variance of the average state prediction for particles with the same sampled model. Uncertainty decomposition allows the user to trade-off aleatoric and epistemic risk \cite{depeweg_decomposition_2018}.} but empirically provides an adequate measure of agent ignorance. For certain scenarios, capturing aleatoric uncertainty is desirable if such environment stochasticity hinders safe control. The dynamics model uncertainty estimation is used as a proxy for risk and penalised in the objective function to promote risk sensitive behaviour. Uncertainty guided CEM planning calculates the return $R_i$ for action sequence $\mathcal{X}_i$ using (\ref{eq:objective}) which results in the planning distribution being updated from an elite-set of action sequences that balance risk and reward.
\begin{equation}
    R_i = \sum_{t=1}^H \left[ r(s_t,a_t) \right] - \beta \omega_i
    \label{eq:objective}
\end{equation}
where $w_i$ is found using (\ref{eq:uncert}) and the hyperparameter $\beta$ controls the trade-off between risk and reward. For $\beta=0$ the algorithm reverts to the original PETS.

\section{Experiments}
\label{sec:experiments}

\begin{figure}[h]
    \centering
    \includegraphics[width=0.75\textwidth]{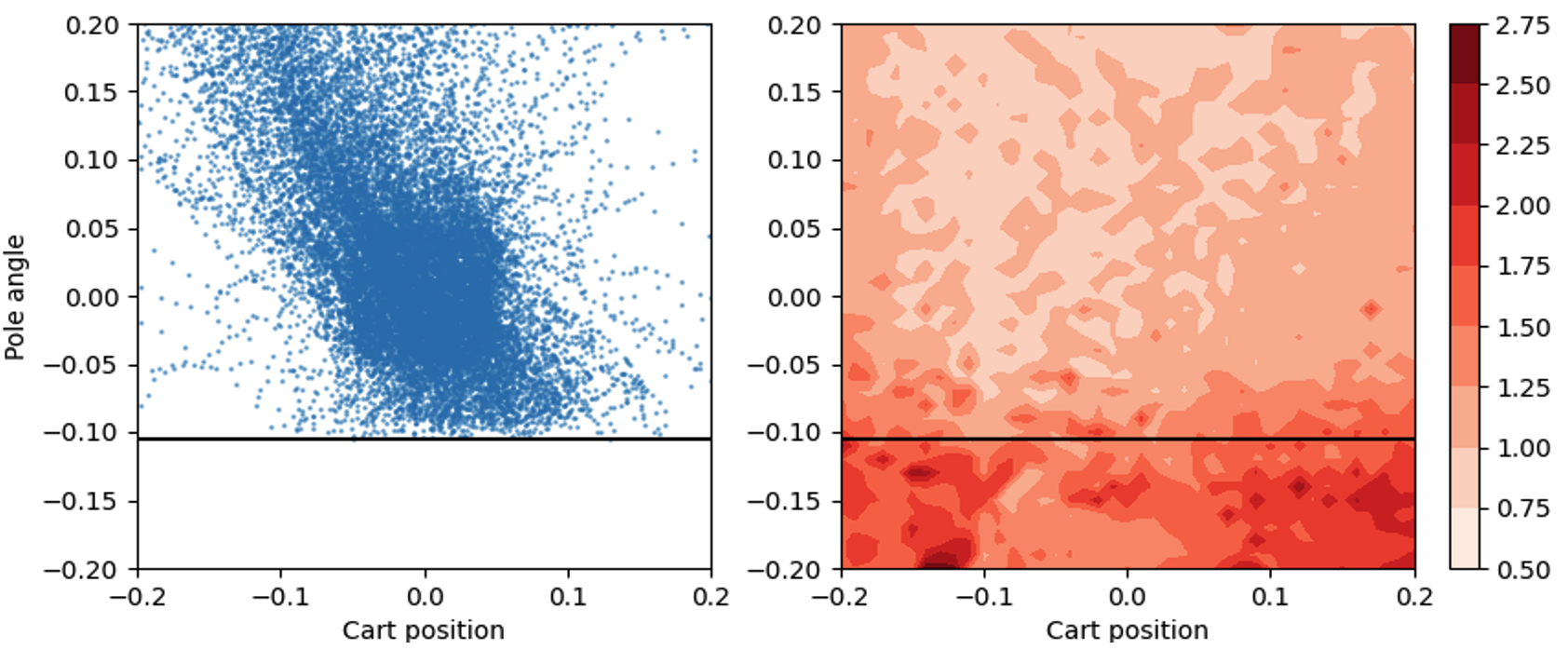}
    \caption{Left: Cartpole training dataset distribution, in which transitions below the line have been removed. Right: Uncertainty predictions for 1-step rollouts ($H=1$) based on the mean variance of a population of 200 actions sampled from the normal distribution.}
    \label{fig:data}
\end{figure}

The objective of the experiments was to test the ability of uncertainty guided CEM planning to identify environment epistemic uncertainty and take mitigating actions to remain in high confidence areas of the state space. This was achieved by training the dynamics models on a subset of the state space with the expectation the agent would identify and avoid out-of-distribution states. An ensemble of probabilistic neural network dynamics models was trained on a dataset collected by random interaction in the Cartpole environment, after removing transitions with a pole angle of $<-0.105\ rad$ ($-6^\circ$). Figure \ref{fig:data} shows the training data distribution and the corresponding estimation of uncertainty, demonstrating out-of-distribution states receive higher measures of uncertainty.

\begin{figure}[b]
    \centering
    \includegraphics[width=\textwidth]{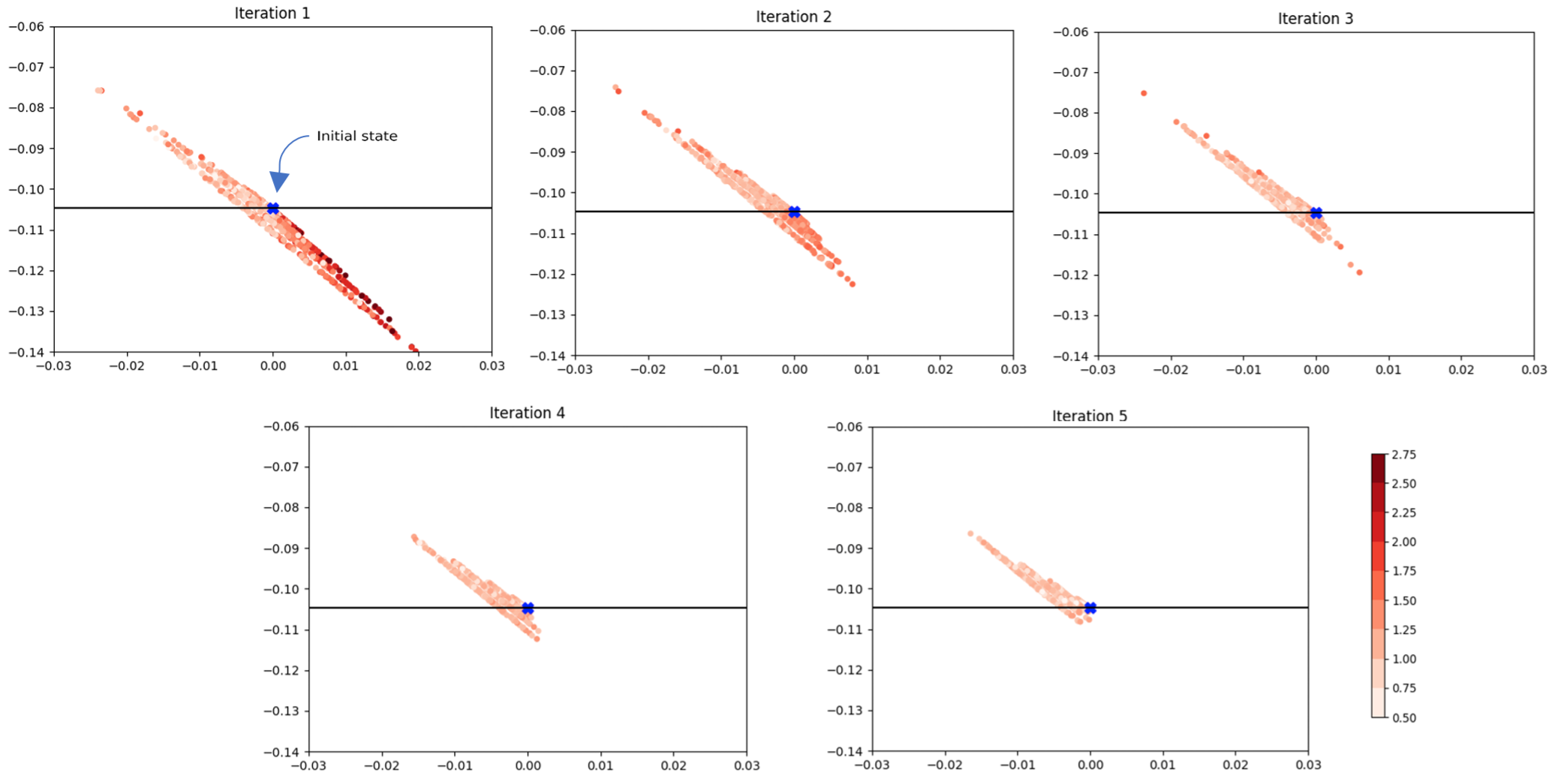}
    \caption{The distribution of states (cart position and pole angle) and uncertainty estimation for 5 planning iterations with a planning horizon $H=5$ from an initial state located on training distribution divide.}
    \label{fig:cem}
\end{figure}

Uncertainty guided CEM planning penalises action sequences in the sampled population that produce high variance state predictions during model rollouts, promoting actions that lead to predictable trajectories. As Figure \ref{fig:cem} shows, this has the effect to skew the planning distribution towards areas of the state space with low uncertainty in the model predictions, which corresponds to regions that are within the training data distribution. As expected, the variance of the distribution of states reached during model rollouts reduces over planning iterations as CEM converges towards the local optimal action sequence. The uncertainty estimation of the state predictions is normalised in (\ref{eq:uncert}) to minimise the effect of compounding variance for states reached later in the planning horizon. The effect of this normalisation can be seen in Figure \ref{fig:cem} as the states with the highest uncertainty are located below the training data divide, rather than at the edges of the distribution of states reached during planning.

\begin{figure}[h]
    \centering
    \includegraphics[width=0.95\textwidth]{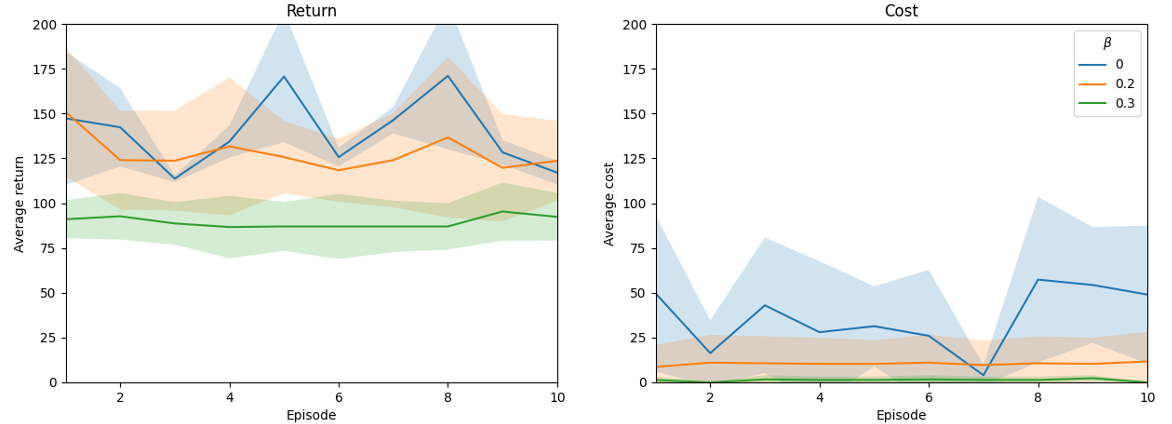}
    \caption{Average return and cost  (number of time steps where pole angle $<-0.105\ rad$) over 3 runs with different random seeds. For experiment hyperparameters, see Appendix \ref{sec:appendix_hyperparams} and for further experiments see Appendix \ref{sec:appendix_experiments}.}
    \label{fig:cartpole}
\end{figure}

The ability of uncertainty guided CEM to perform safe planning was tested by running episodes in an offline scenario, whereby the dynamics models are fixed after training on the original dataset. As a result of no further online exploration or model learning, the performance of the agent is limited and summarised as an average across all episodes in the experiment. Figure \ref{fig:cartpole} shows the ability of uncertainty guided CEM planning to reduce the time spent in the out-of-distribution area of the state space\footnote{The cost tracked is simply a sum of time steps spent in the out-of-distribution area of the state-space and is not available to the agent during optimisation. This is opposed to a cost associated with a constrained MDP \cite{altman_constrained_1999}.}, without enforcing an explicit hard constraint. The consequence is a reduction to the reward attained, where the hyperparameter $\beta$ controls the trade-off between risk and reward. Uncertainty guided planning also resulted in a reduction to the variability of return and cost over a number of runs with different random seeds displaying more stable behaviour.


\section{Discussion and Conclusion}
\label{sec:conclusion}
 Experiments were conducted in an offline setting, which eliminates the need for exploration and isolates the planning algorithm from learning, allowing analysis of uncertainty estimation and risk-sensitive optimisation. The performance of offline RL algorithms are fundamentally limited to the size and diversity of the training dataset. To improve performance, online exploration of the environment is required to reduce dynamics model uncertainty and expand the known region of the state space. For online learning, the hyperparameter $\beta$ in uncertainty guided CEM determines the extent of exploration and could be set based on an acceptable level of risk. Alternative safe exploration strategies include those that draw on human knowledge through advice or intervention \cite{saunders_trial_2018}, methods based on constraint satisfaction during active learning \cite{cowen-rivers_samba_2020}, or by guaranteeing Lyapunov stability to expand the region of attraction \cite{berkenkamp_safe_2017}.

The goal of safe RL is to avoid dangerous situations as defined by implicit or explicit constraints, whilst offline MBRL aims to mitigate model exploitation due to distribution shift in both the state and action distributions. As shown here, and by recently proposed MBRL algorithms \cite{yu_mopo_2020,zhang_cautious_2020}, one approach to address both these problems uses dynamics model uncertainty estimation to identify unknown regions of the environment. Furthermore, offline pre-training is desirable in safety-critical domains where online learning poses unacceptable risk. Future research will focus on the intercept of offline RL and safe learning, focusing on problems with predefined constraints.

Uncertainty guided CEM planning has been shown to avoid unknown, and thus potentially dangerous, regions of the state space by identifying epistemic uncertainty through the use of an ensemble of dynamics models, and taking mitigating actions to remain in high confidence areas. In safety-critical scenarios such behaviour is desirable as low confidence exploratory actions could have drastic consequences.

\newpage
\begin{ack}
We would like to thank Tom Bewley and Jonathan Thomas for their useful discussions and feedback while conceptualising this paper.

Funding in direct support of this work: EPSRC Centre for Doctoral Training Studentship (EP/S022937/1).

The authors declare no conflicts of interest.
\end{ack}

{\small
\bibliographystyle{ieeetr}
\bibliography{references}
}

\newpage
\appendix

\section{Appendix}
\label{sec:appendix}

\subsection{Experiment Details}
\label{sec:appendix_hyperparams}

MBRL-Lib (MIT Licence) \cite{pineda_mbrl-lib_2021} was used for experiments using the Cartpole environment with continuous actions with the following hyperparameters.

\begin{table}[h]
  \caption{Hyperparameters}
  \label{experiment_details}
  \centering
  \begin{tabular}{ll}
    \toprule
    \multicolumn{2}{c}{Dynamics model and training} \\
    \cmidrule(r){1-2}
    Number of layers & 3 \\
    Nodes in hidden layers & 200 \\
    Activation function & ReLU \\
    Ensemble size ($B$) & 4 \\
    Replay buffer size ($D$) & 10000 \\
    Training batch size & 32 \\
    Epochs & 50 \\
    Optimiser & Adam \\
    Learning rate & 0.001 \\ 
    Weight decay & 0.00005 \\
    \midrule
    \multicolumn{2}{c}{CEM planning} \\
    \cmidrule(r){1-2}
    Horizon ($H$) & 10 \\
    Planning iterations & 5 \\
    Elite ratio & 0.3 \\
    Population size ($N$) & 200 \\
    Number of particles ($P$) & 12 \\
    Alpha & 0.1 \\

    \bottomrule
  \end{tabular}
\end{table}

\subsection{Further Experiments}
\label{sec:appendix_experiments}

The dynamics models were trained on a dataset collected by random interaction with the Pendulum environment after transitions with a pendulum angle $-\frac{3}{4}\pi < \theta < -\frac{1}{4}\pi$ were removed. The episodes were run for 200 time steps in an offline setting.

\begin{figure}[h]
    \centering
    \includegraphics[width=\textwidth]{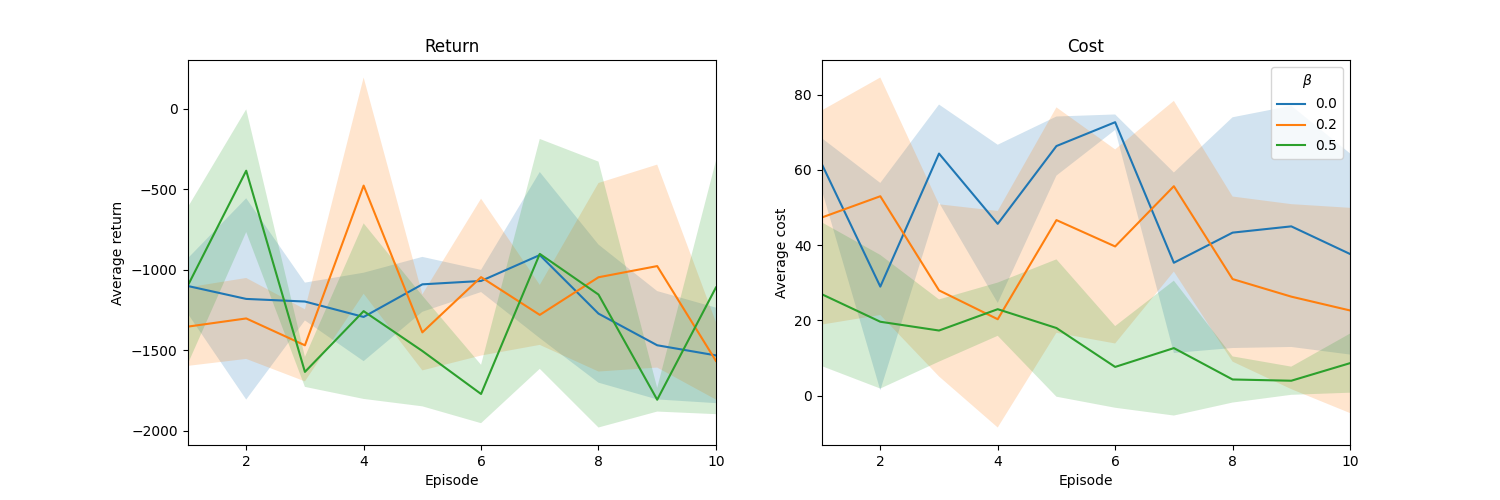}
    \caption{Average return and cost (number of time steps where pendulum angle $-\frac{3}{4}\pi < \theta < -\frac{1}{4}\pi$) over 3 runs with different random seeds.}
\end{figure}

\end{document}